\documentclass[10pt,twocolumn,letterpaper]{article}

\usepackage[pagenumbers]{cvpr} 

\usepackage{graphicx}
\usepackage{amsmath}
\usepackage{amssymb}
\usepackage{booktabs}
\usepackage{tabularx}
\usepackage{algorithm}
\usepackage{algpseudocode}
\usepackage{xfrac}
\usepackage{multirow}
\usepackage[accsupp]{axessibility}  

\usepackage[pagebackref,breaklinks,colorlinks]{hyperref}

\usepackage[capitalize]{cleveref}
\crefname{section}{Sec.}{Secs.}
\Crefname{section}{Section}{Sections}
\Crefname{table}{Table}{Tables}
\crefname{table}{Tab.}{Tabs.}

\newcommand{\minisection}[1]{\vspace{2mm}\noindent{\textbf{#1}}}
\newcommand{\anh}[1]{\textcolor{cyan}{[#1]}}

\def\cvprPaperID{10360} 
\def\confName{CVPR}
\def\confYear{2023}

\begin{document}

\title{Wavelet Diffusion Models are fast and scalable Image Generators}

\author{
Hao Phung, Quan Dao, Anh Tran\\
VinAI Research\\
{\tt\small \{v.haopt12, v.quandm7, v.anhtt152\}@vinai.io}
}

\maketitle

\begin{abstract}
    Diffusion models are rising as a powerful solution for high-fidelity image generation, which exceeds GANs in quality in many circumstances. However, their slow training and inference speed is a huge bottleneck, blocking them from being used in real-time applications. A recent DiffusionGAN method significantly decreases the models' running time by reducing the number of sampling steps from thousands to several, but their speeds still largely lag behind the GAN counterparts. This paper aims to reduce the speed gap by proposing a novel wavelet-based diffusion scheme. We extract low-and-high frequency components from both image and feature levels via wavelet decomposition and adaptively handle these components for faster processing while maintaining good generation quality. Furthermore, we propose to use a reconstruction term, which effectively boosts the model training convergence. Experimental results on CelebA-HQ, CIFAR-10, LSUN-Church, and STL-10 datasets prove our solution is a stepping-stone to offering real-time and high-fidelity diffusion models. Our code and pre-trained checkpoints are available at \url{https://github.com/VinAIResearch/WaveDiff.git}.
\end{abstract}

\section{Introduction} \label{sec:intro}
\begin{figure}[t]
  \centering
   \includegraphics[width=\linewidth]{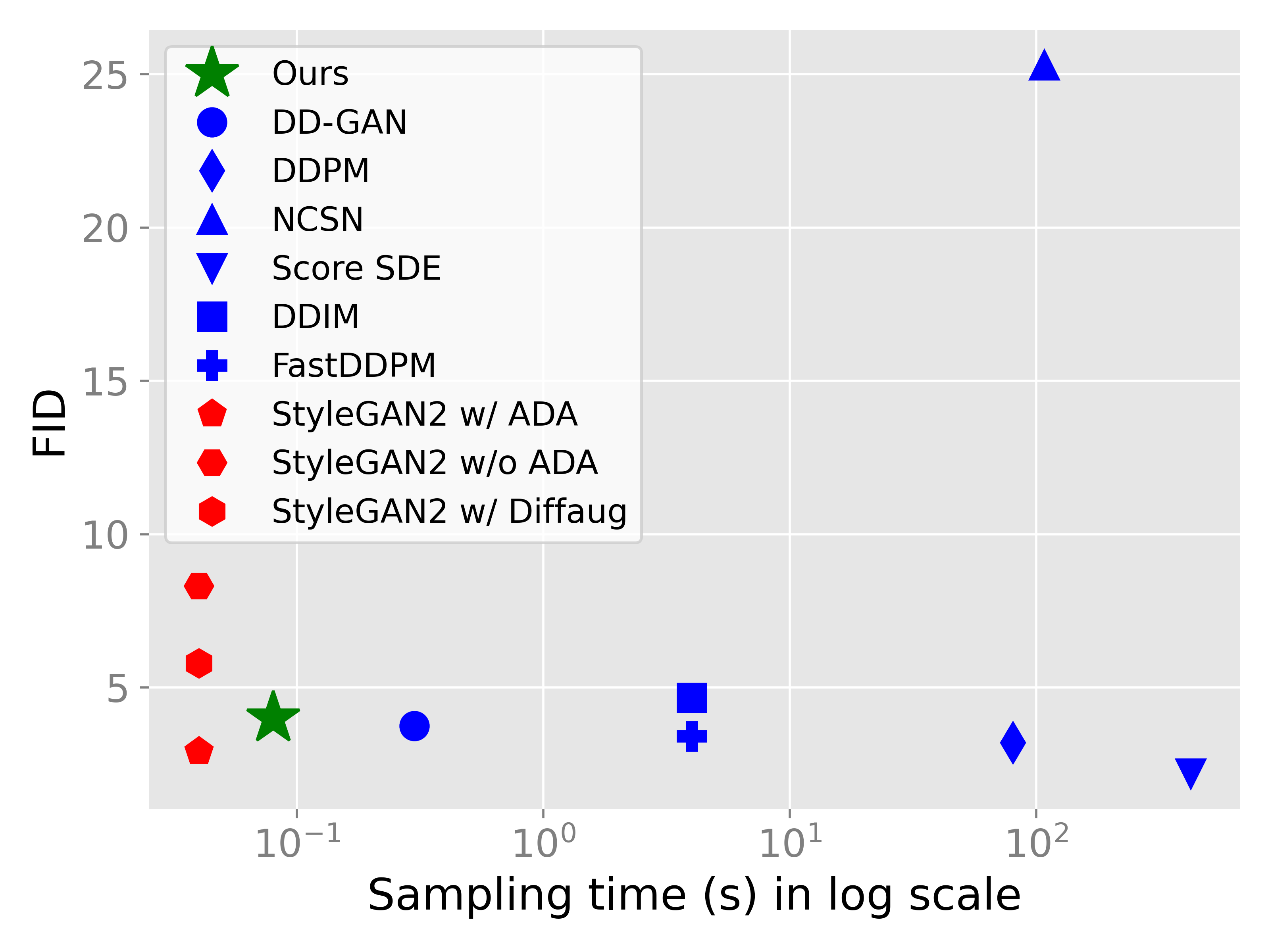}
   \vspace{-7mm}
   \caption{Comparisons between our method and other GAN and diffusion techniques in terms of FID and sampling time the on CIFAR-10 ($32\times 32$) dataset. Our method is $2.5\times$ faster than DDGAN~\cite{xiao2021tackling}, the fastest up-to-date diffusion method, and approaches the real-time speed of StyleGAN methods~\cite{karras2019style,karras2020training,zhao2020differentiable} while still achieving comparable FID scores.}
   \label{fig:teaser}
   \vspace{-2mm}
\end{figure}
Despite being introduced recently, diffusion models have grown tremendously and drawn many research interests. Such models revert the diffusion process to generate clean, high-quality outputs from random noise inputs. These techniques are applied in various data domains and applications but show the most remarkable success in image-generation tasks. Diffusion models can beat the state-of-the-art generative adversarial networks (GANs) in generation quality on various datasets \cite{dhariwal2021diffusion,saharia2022photorealistic}. More notably, diffusion models provide a better mode coverage \cite{song2021maximum,kingma2021variational,huang2021variational} and a flexible way to handle different types of conditional inputs such as semantic maps, text, representations, and images \cite{rombach2022high}. Thanks to this capability, they offer various applications such as text-to-image generation, image-to-image translation, image inpainting, image restoration, and more. Recent diffusion-based text-to-image generative models \cite{ramesh2022hierarchical,BibEntry2022Nov,saharia2022photorealistic} allow users to generate unbelievably realistic images just by text inputs, opening a new era of AI-based digital art and promising applications to various other domains.

While showing great potential, diffusion models have a very slow running speed, a critical weakness blocking them from being widely adopted like GANs. The foundation work Denoising Diffusion Probabilistic Models (DDPMs) \cite{ho2020denoising} requires a thousand sampling steps to produce the desired output quality, taking minutes to generate a single image. Many techniques have been proposed to reduce the inference time \cite{song2020denoising,kong2021fast}, mainly via reducing the sampling steps. However, the fastest algorithm before DiffusionGAN still takes seconds to produce a 32$\times$32 image, which is about 100 times slower than GAN. DiffusionGAN \cite{xiao2021tackling} made a break-though in fastening inference speed by combining Diffusion and GANs in a single system, which ultimately reduces the sampling steps to 4 and the inference time to generate a 32$\times$32 image as a fraction of a second. It makes DiffusionGAN the fastest existing diffusion model. Still, it is at least 4 times slower than the StyleGAN counterpart, and the speed gap consistently grows when increasing the output resolution. Moreover, DiffusionGAN still requires a long training time and a slow convergence, confirming that diffusion models are not yet ready for large-scale or real-time applications.

This paper aims to bridge the speed gap by introducing a novel wavelet-based diffusion scheme. Our solution relies on discrete wavelet transform, which decomposes each input into four sub-bands for low- (LL) and high-frequency (LH, HL, HH) components. We apply that transform on both image and feature levels. This allows us to significantly reduce both training and inference times while keeping the output quality relatively unchanged. On the image level, we obtain a high speed boost by reducing the spatial resolution four times. On the feature level, we stress the importance of wavelet information on different blocks of the generator. With such a design, we can obtain considerable performance improvement while inducing only a marginal computing overhead.  

Our proposed Wavelet Diffusion provides state-of-the-art training and inference speed while maintaining high generative quality, thoroughly confirmed via experiments on standard benchmarks including CIFAR-10, STL-10, CelebA-HQ, and LSUN-Church. Our models significantly reduce the speed gap between diffusion models and GANs, targeting large-scale and real-time systems.

In summary, our contributions are as following:
\begin{itemize}
    \item We propose a novel Wavelet Diffusion framework that takes advantage of the dimensional reduction of Wavelet subbands to accelerate Diffusion Models while maintaining good visual quality of generated results through high-frequency components.
    \item We employ wavelet decomposition in both image and feature space to improve generative models' robustness and execution speed. 
    \item Our proposed Wavelet Diffusion provides state-of-the-art training and inference speed, which serves as a stepping-stone to facilitating real-time and high-fidelity diffusion models. 
\end{itemize}


\section{Related work}
\label{sec:rel_work}

\subsection{Diffusion models} 

Diffusion models~\cite{sohl2015deep,ho2020denoising} are inspired by non-equilibrium thermodynamics where the diffusion and reverse processes are Markov chains. Unlike GANs~\cite{goodfellow2014generative,karras2019style} and VAEs~\cite{kingma2014vae,vahdat2020nvae}, they sample using a large number of denoising steps with a fixed procedure where each latent variable shares the same dimensionality as the original inputs. A line of methods shares the same motives but relies on score matching for the reverse process~\cite{song2019generative,song2020score,vincent2011connection}. Later works focus on improving sample quality of diffusion models~\cite{song2020improved,dhariwal2021diffusion,song2021maximum}. Some have shifted the process to latent space~\cite{rombach2022high,vahdat2021score}, which enables the success of text-to-image generation~\cite{ramesh2021zero,esser2021imagebart,saharia2022photorealistic,ramesh2022hierarchical}. Similar to us, some approaches~\cite{ma2022accelerating,luhman2021knowledge} aim to improve the efficiency of sampling process for better convergence and faster sampling. As the Markov process is a sequential probability model with a finite step, some try to break it into non-Markov chains for faster sampling~\cite{song2020denoising}. Despite several efforts to accelerate the sampling process, there is an inevitable trade-off between sampling speed and quality. 


\minisection{Diffusion GAN \cite{xiao2021tackling}} was recently proposed as a novel approach to sidestep the unimodal Gaussian assumption of small step sizes via modeling the complex multimodal distribution of large step sizes with generative adversarial networks. This proposal greatly reduces the number of denoising steps to just a finger count (e.g., 2, 4), which leads the inference time to a fraction of a second. However, it is still largely slower than GAN competitors. Hence, we further maximize its full potential by introducing new wavelet components on top of this framework.
\subsection{Wavelet-based approaches} Wavelet decomposition \cite{mallat1989theory,chui1992introduction} is a classical operator widely used in various computer vision tasks. It is the fundamental process behind a popular JPEG-2000 image compression format \cite{Taubman2012}. In recent years, wavelet transform has started being incorporated into many deep-learning-based systems as they can take advantage of spatial and frequency information to aid the training. Some have applied it to the design of networks to improve visual representation learning \cite{li2020wavelet,liu2019multi,yao2022wave,jiang2021focal}. Some have extended it to image restoration~\cite{yu2021wavefill, deng2019wavelet}, style transfer~\cite{yoo2019photorealistic}, and face problems~\cite{gao2021high,huang2019wavelet}. On the line of image generation, wavelet transform is plugged in recent generative adversarial networks~\cite{Yang2022WaveGAN, zhang2022styleswin, gal2021swagan,yang2022fregan} to improve the visual quality of output images. 

Regarding diffusion model families, several works start employing wavelet transform. Guth et al. \cite{guth2022wavelet} accelerate score-based models by modeling conditional probabilities of multiscale wavelet coefficients, but still suffer from low-quality output images. Li et al. \cite{li2022wavelet} perform score matching on the wavelet domain for better image colorization.

However, none of the mentioned techniques can balance the diffusion models' quality and speed. Here, we introduce a novel wavelet diffusion approach to not only emphasize the importance of frequency information but also to reduce the sampling time by a large margin. We take advantage of frequency sparsity and dimensionality reduction of wavelet transform to reduce high-dimensional space to the lower-dimensional manifolds, which are essential for faster and more efficient sampling. Unlike existing methods on diffusion families, we enrich the frequency awareness on both image and feature levels to facilitate high-fidelity image generation. This is greatly inspired by the success of wavelet-based GANs.

\section{Background}
\label{sec:background}

\subsection{Diffusion Models}
The traditional diffusion process requires thousand timesteps to gradually diffuse each input $x_0$, following a data distribution $p(x_0)$, into pure Gaussian noise. The posterior probability of a diffused image $x_t$ at timestep $t$ has a closed form:
\begin{equation}
    q(x_t|x_0) = \mathcal{N}(x_t; \sqrt{\bar{\alpha_t}}x_0, (1 - \bar{\alpha_t})\mathbf{I})
\end{equation}
where $\alpha_t = 1 - \beta_t$, $\bar{\alpha_t} = \prod_{s=1}^t \alpha_s$, and $\beta_t \in (0,1)$ is defined to be small through a variance schedule which can be learnable or fixed at timestep $t$ in the forward process. 

Since the diffusion process adds relatively small noise per step, the reverse process $q(x_{t-1}|x_t)$ can be approximated by Gaussian process $q(x_{t-1}|x_t,x_0)$. Therefore, the trained denoising process $p_\theta(x_{t-1}|x_t)$ can be parameterized according to $q(x_{t-1}|x_t,x_0)$. The common parameterized form of $p_\theta(x_{t-1}|x_t)$ is:
\begin{equation}
    p_\theta(x_{t-1}\mid x_t) = \mathcal{N}(x_{t-1}; \mu_\theta(x_t, t), \sigma^2_t\mathbf{I}),
\end{equation}
where $\mu_\theta(x_t, t)$ and $\sigma^2_t$  are the mean and variance of the parametric denoising model, respectively. The objective is to minimize the distance between a true denoising distribution $q(x_{t-1}|x_t)$ and the parameterized one $p_\theta(x_{t-1}|x_t)$ through Kullback-Leibler (KL) divergence.


 Unlike the traditional diffusion methods, DiffusionGAN~\cite{xiao2021tackling} enables large step sizes for faster sampling through generative adversarial networks. It introduces a discriminator $D_\phi$ and optimizes both the generator and the discriminator in an adversarial training manner:
\begin{multline}
    \min _\phi \max _\theta \sum_{t \geq 1} \mathbb{E}_{q\left(\mathbf{x}_t\right)}[\mathbb{E}_{q\left(\mathbf{x}_{t-1} \mid \mathbf{x}_t\right)}[-\log (D_\phi(\mathbf{x}_{t-1}, \mathbf{x}_t, t))] \\
    + \mathbb{E}_{p_\theta\left(\mathbf{x}_{t-1} \mid \mathbf{x}_t\right)}[\log (D_\phi(\mathbf{x}_{t-1}, \mathbf{x}_t, t))]],
    \label{eq:ddgan_obj}
\end{multline}
where fake samples are sampled from a conditional generator $p_\theta\left(\mathbf{x}_{t-1} \mid \mathbf{x}_t\right)$. As large step sizes cause $q(x_{t-1}|x_t)$ to be no longer a Gaussian distribution, DiffusionGAN~\cite{xiao2021tackling} aims to implicitly model this complex multimodal distribution with a generator $G_\theta(x_t, z, t)$ given a $D$-dimensional latent variable $z \sim \mathcal{N}(0, \mathbf{I})$.
Specifically, DiffusionGAN first generates unperturbed sample $x_0'$ through the generator $G_\theta(x_t, z, t)$ and acquires the corresponding perturbed sample $x_{t-1}'$ using $q(x_{t-1}|x_t,x_0)$. Meanwhile, the discriminator performs judgment on real pairs $D_\phi(x_{t-1}, x_t,t)$ and fake pairs $D_\phi(x_{t-1}', x_t,t)$.

For convenience, we will abbreviate DiffusionGAN as DDGAN in later sections.




 
\subsection{Wavelet Transform}

Wavelet Transform is a classical technique widely used in image compression to separate the low-frequency approximation and the high-frequency details from the original image. While low subbands are similar to down-sampled versions of the original image, high subbands express the local statistics of vertical, horizontal, and diagonal edges. Notably, the Haar wavelet is widely adopted in real-world applications due to its simplicity. It involves two types of operations: discrete wavelet transform (DWT) and discrete inverse wavelet transform (IWT). 

Let denote $L = \frac{1}{\sqrt{2}}\begin{bmatrix}
    1 & 1\end{bmatrix}$ and $H = \frac{1}{\sqrt{2}}\begin{bmatrix} -1 & 1\end{bmatrix}$ be low-pass and high-pass filters. They are used for constructing four kernels with stride 2, namely $LL^T, LH^T, HL^T$, and $HH^T$  to decompose the input $X \in \mathbb{R}^{H\times W}$ into four subbands $X_{ll}, X_{lh}, X_{hl}$, and $X_{hh}$ with a size of $H/2\times W/2$. As these filters are pairwise orthogonal, they can form a $4\times4$ invertible matrix to accurately reconstruct original signals $X$ from frequency components by IWT.


In this paper, we use this kind of transform to decompose input images and feature maps to emphasize high-frequency components and reduce the spatial dimensions to four folds for more efficient sampling.


\section{Method}\label{sec:method}
This section describes our proposed Wavelet Diffusion framework. First, we present the core wavelet-based diffusion scheme for more efficient sampling (\cref{sec:wavelet_diffusion}). We then depict the design of a new wavelet-embedded generator for better frequency-aware image generation (\cref{sec:wavelet_feature}). 

\subsection{Wavelet-based diffusion scheme}\label{sec:wavelet_diffusion}
\begin{figure*}[t]
  \centering
   \includegraphics[width=0.8\linewidth]{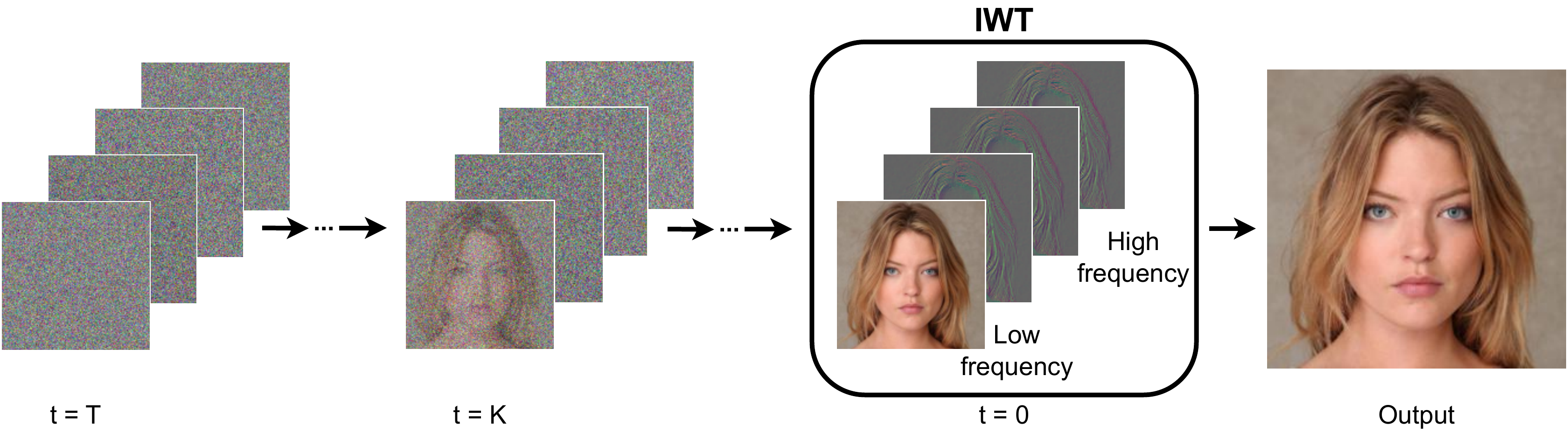}
   \vspace{-3mm}
   \caption{Illustration of Wavelet-based diffusion scheme. It performs denoising on wavelet space instead of pixel space. At each step $t$, a less-noisy sample $y_{t-1}$ is generated by a denoiser $p_\theta(y_{t-1} \vert y_{t})$ with parameters $\theta$. After obtaining the clean sample $y_0$ through $T$ steps, it is used to reconstruct the final image via Inverse wavelet transformation (IWT).}
   \label{fig:single}
\end{figure*}

First, we describe how to incorporate wavelet transform in the diffusion process. We decompose the input image into four wavelet subbands and concatenate them as a single target for the denoising process (illustrated in~\cref{fig:single}). Such a model does not perform on the original image space but on the wavelet spectrum. As a result, our model can leverage high-frequency information to increase the details of generated images further. Meanwhile, the spatial area of wavelet subbands is four times smaller than the original image, so the computational complexity of the sampling process is significantly reduced. 

We build our method upon DDGAN model where the input is 4 wavelet subbands of wavelet transformation. Given input image $x \in \mathbb{R}^{3 \times H \times W}$, we decompose it into a set of low and high subbands which are further concatenated to form a matrix $y \in \mathbb{R}^{12\times \frac{H}{2} \times \frac{W}{2}}$. This input is then \textbf{\textit{projected to the base channel D}} via the first linear layer, keeping the network width unchanged compared with DDGAN. Hence, most of the network benefits from $4\times$ reduction in spatial dimensions, significantly reducing its computation.

Let denote $y_0$ be a clean sample and $y_t$ be a corrupted sample at timestep $t$ which is sampled from $q(y_t | y_0)$. In terms of the denoising process, a generator receives a tuple of variable $y_t$, a latent $z \sim \mathcal{N}(0, \mathbf{I})$, and a timestep $t$ to generate an approximation of original signals $y_0$: $y_0' = G(y_t, z, t)$. The predicted noisy sample $y'_{t-1}$ is then drawn from tractable posterior distribution $q(y_{t-1}|y_t, y'_0)$. The role of the discriminator is to distinguish the real pairs $(y_{t-1}, y_t)$ and the fake pairs $(y'_{t-1}, y_t)$.

\begin{algorithm}[t]
\caption{Wavelet-based sampling process}
\label{alg:singlescale_sample}
\centering
\small
\begin{algorithmic}
\State $y_t \sim \mathcal{N}(0, \mathbf{I})$

\For{$t = T,\cdots,1$}
    \State $z \sim \mathcal{N}(0, \mathbf{I})$
    \State $y_0' = G(y_t, z, t)$
    \State $y_{t-1} \sim q(y_{t-1}|y_t, y_0')$
\EndFor
\State $x_0 = \text{IWT}(y_0)$
\State \textbf{return} $x_0$
\end{algorithmic}
\end{algorithm}

\minisection{Adversarial objective} Following\cite{xiao2021tackling}, we optimize the generator and the discriminator through the adversarial loss:
\begin{equation}
    \label{eq:adv_loss}
    \begin{aligned}
        \mathcal{L}_{adv}^D &= -\log(D(y_{t-1}, y_t, t)) + \log(D(y'_{t-1}, y_t, t)), \\
        \mathcal{L}_{adv}^G &= -\log(D(y'_{t-1}, y_t, t)).
    \end{aligned}
\end{equation}

\minisection{Reconstruction term} In addition to the adversarial objective in ~\cref{eq:adv_loss}, we add a reconstruction term to not only impede the loss of frequency information but also preserve the consistency of wavelet subbands. It is formulated as an L1 loss between a generated image and its ground-truth:
\begin{equation}
    \label{eq:rec_loss}
    \mathcal{L}_{rec} = \Vert y_0' - y_0 \Vert
    .
\end{equation}

The overall objective of the generator is a linear combination of adversarial loss and reconstruction loss:
\begin{equation}
    \mathcal{L}^G = \mathcal{L}_{adv}^G + \lambda \mathcal{L}_{rec},
\end{equation}
where $\lambda$ is a weighting hyper-parameter (default value is 1). 

After a few sampling steps as defined, we acquire the estimated denoised subbands $y_0'$. The final image can be recovered via wavelet inverse transformation $x_0' = \text{IWT}(y_0')$. We depict the sampling process in~\cref{alg:singlescale_sample}. 

\subsection{Wavelet-embedded networks}\label{sec:wavelet_feature}
Next, we further incorporate wavelet information into feature space through the generator to strengthen the awareness of high-frequency components. This is beneficial to the sharpness and quality of final images.

\begin{figure*}[t]
  \centering
   \includegraphics[width=0.7\linewidth]{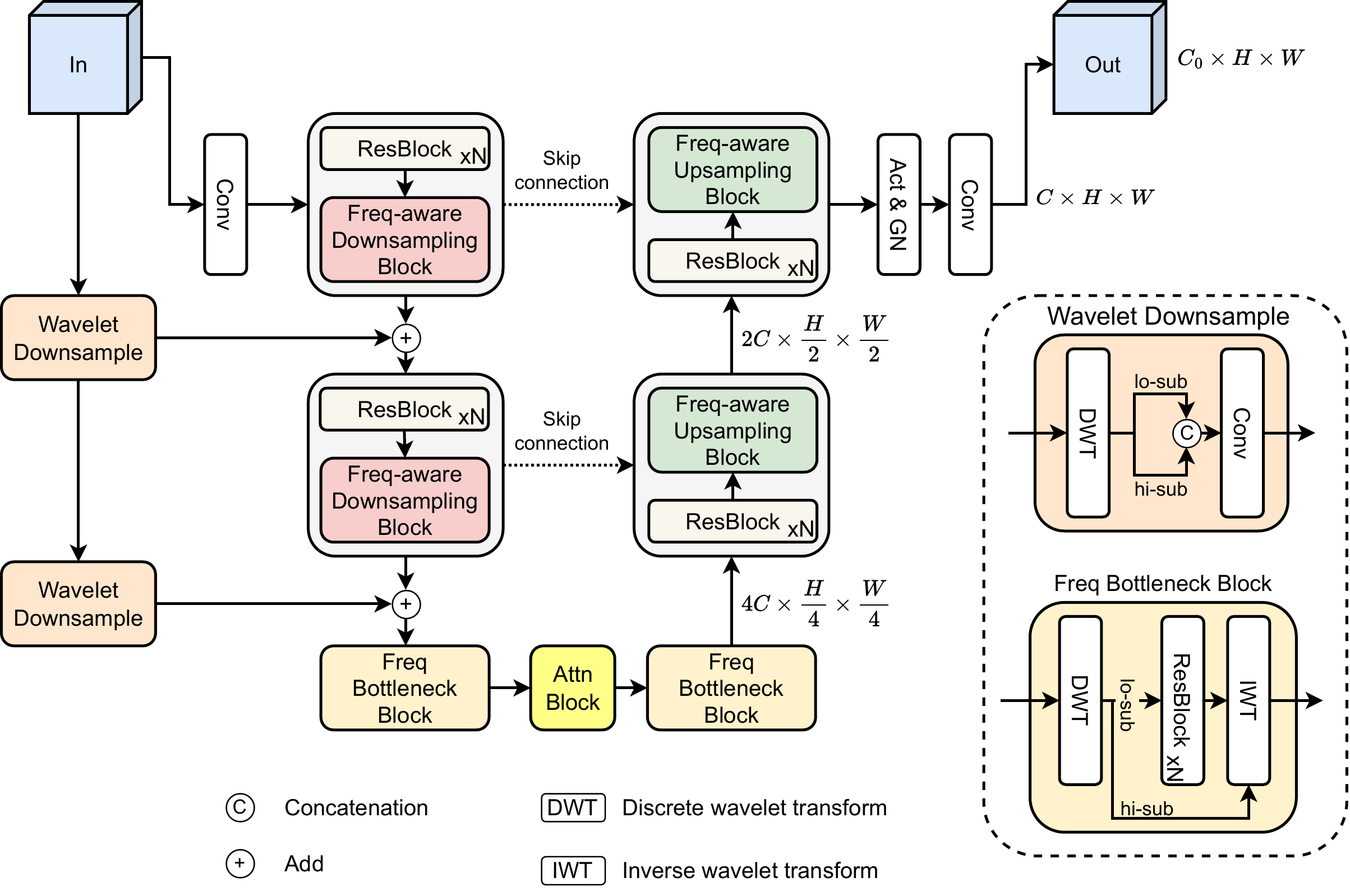}
   \vspace{-2mm}
   \caption{Illustration of Wavelet-embedded generator. For simplification, timestep embedding $t$ and latent embedding $z$ are ignored but they are injected in individual blocks of the denoising process. The inputs are noisy wavelet subbands of shape $[12 \times H \times W]$ at timestep $t$, which are processed by a sequence of our proposed components, including frequency-aware upsampling and downsampling blocks, frequency residual connections, and a brand new frequency bottleneck block. The outputs of the model are the approximation of unperturbed inputs.}
   \label{fig:wunet}
   \vspace{-2mm}
\end{figure*}

\cref{fig:wunet} illustrates the structure of our proposed wavelet-embedded generator. It follows the UNet structure of \cite{song2020score} with $M$ down-sampling and $M$ up-sampling blocks plus skip connections between blocks of the same resolution, with $M$ predefined. However, instead of using the normal downsampling and upsampling operators, we replace them with frequency-aware blocks. At the lowest resolution, we employ frequency-bottleneck blocks for better attention on low and high-frequency components. Finally, to incorporate original signals $Y$ to different feature pyramids of the encoder, we introduce frequency residual connections using wavelet downsample layers.
Let denote $Y$ be the input image and $F_i$ is the $i$-th intermediate feature map of $Y$. We will discuss below the newly introduced components:

\begin{figure}[t]
  \centering
   \includegraphics[width=0.8\linewidth]{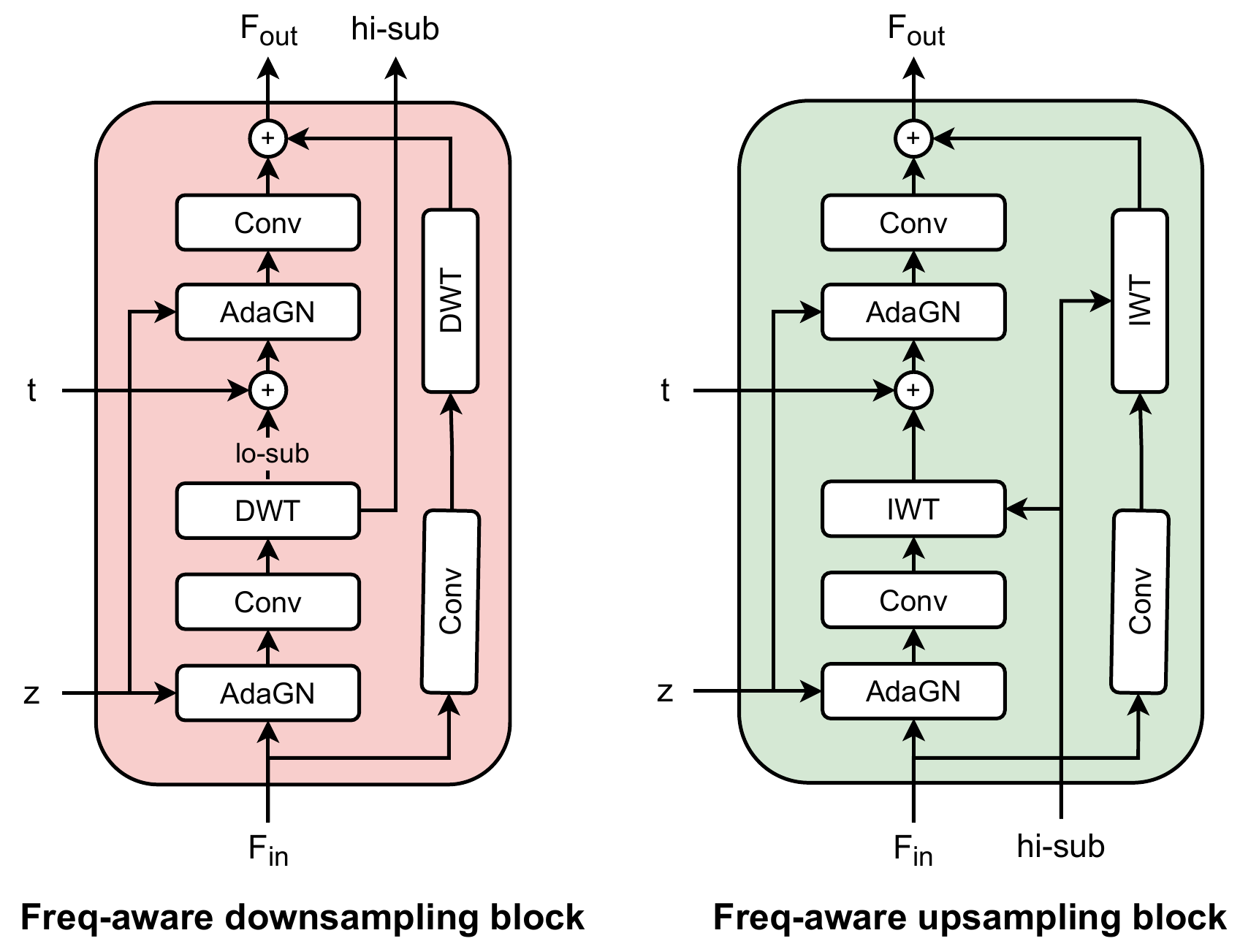}
   \vspace{-2mm}
   \caption{Details of frequency-aware downsampling and upsampling blocks in the wavelet-embedded generator.}
   \label{fig:wunet_blocks}
   \vspace{-3mm}
\end{figure}

\minisection{Frequency-aware downsampling and upsampling blocks.} Traditional approaches relied on a blurring kernel for the downsampling and upsampling process to mitigate the aliasing artifact. We instead utilize inherent properties of the wavelet transform for better upsampling and downsampling (depicted in~\cref{fig:wunet_blocks}). This, in fact, strengthens the awareness of high-frequency information on these operations. Particularly, the downsampling block receives a tuple of input features $F_i$, a latent $z$, and time embedding $t$, which are then processed through a sequence of layers to return downsampled features and high-frequency subbands. These returned subbands are served as an additional input to upsample features based on frequency cues in the upsampling block. 

\minisection{Frequency bottleneck block} locates at the middle stage, which includes two frequency bottleneck blocks and one attention block in-between. Each frequency bottleneck block first divides feature map $F_i$ into the low-frequency subband $F_{i, ll}$ and the concatenation of high-frequency subbands $F_{i, H}$. $F_{i, ll}$ is then passed as input to resnet block(s) for deeper processing. The processed low-frequency feature map and the original high-frequency subbands $F_{i, H}$ are transformed back to the original space via IWT.  With such a bottleneck, the model can focus on learning intermediate feature representations of low-frequency subbands while preserving the high-frequency details.

\minisection{Frequency residual connection} The original design of the network in \cite{song2020score} incorporates original signals $Y$ to different feature pyramids of the encoder via a strided-convolution downsampling layer. We instead use a wavelet downsample layer to map residual shortcuts of input $Y$ to the corresponding feature dimensions, which are then added to each feature pyramid. Specifically, the residual shortcuts of $Y$ are decomposed into four subbands which are then concatenated and fed to a convolution layer for feature projection. This shortcut aims to enrich the perception of the frequency source of feature embeddings.

\begin{table*}[t]
    \centering
        \begin{tabular}{l|lll|lll}
        \toprule
        \multirow{2}{*}{}       & \multicolumn{3}{c|}{\textbf{DDGAN}} & \multicolumn{3}{c}{\textbf{Ours}} \\
                                & Params & FLOPs   & MEM    & Params  & FLOPs  & MEM   \\
        \midrule
        CIFAR-10                & 48.43M & 7.05G   & 0.31G  & 33.37M  & 1.67G  & 0.16G \\
        STL-10                  & 48.43M & 27.26G  & 0.66G  & 55.58M  & 7.15G  & 0.34G \\
        LSUN \& CelebA HQ (256) & 39.73M & 70.82G  & 3.21G  & 31.48M  & 28.54G & 1.07G \\
        CelebA HQ (512)         & 39.73M  & 282.00G & 12.30G & 44.85M   & 74.35G & 3.22G \\
        \bottomrule
        \end{tabular}
    \vspace{-2mm}
    \caption{Model specifications of DDGAN\cite{xiao2021tackling} and our approach including a number of parameters (M), FLOPs (GB), and memory usage (GB) on a single GPU for one sample. Our method attains lower computing FLOPs and memory consumption while having a comparable number of parameters.}
    \vspace{-2mm}
    \label{tab:model_specs}
\end{table*}

\section{Experiments} \label{sec:exp}
In this section, we first provide details about the experimental setup and then present the empirical experiments on different datasets, including CIFAR-10, STL-10, CelebA-HQ, and LSUN. Finally, we ablate the important components of our proposed framework.
\subsection{Experimental setup} 
\minisection{Datasets} We experiment on CIFAR-10 $32\times32$, STL-10 $64\times64$ and two other higher-resolution datasets, including CelebA-HQ $256\times256$ and LSUN-Church $256\times256$. We also examine our model training on high-resolution images with CelebA-HQ (512 \& 1024).

\begin{table}[t]
  \centering
  \resizebox{\linewidth}{!}{

\addtolength{\tabcolsep}{-1pt}   
  \begin{tabular}{lcccc}
    \toprule
    Model & FID$\downarrow$ & Recall$\uparrow$ & NFE$\downarrow$ & Time (s)$\downarrow$ \\
    \midrule
    Ours & 4.01 & 0.55 & 4 & \textbf{0.08} \\
    DDGAN \cite{xiao2021tackling} & \textbf{3.75} & 0.57 & 4 & 0.21 (0.30$^*$) \\
    \midrule
    DDPM \cite{ho2020denoising} & 3.21 & 0.57 & 1000 & 80.5 \\
    NCSN \cite{song2019generative} & 25.3 & - & 1000 & 107.9\\
    Score SDE (VE) \cite{song2020score} & 2.20 & 0.59 & 2000 & 423.2 \\
    Score SDE (VP) \cite{song2020score} & 2.41 & 0.59 & 2000 & 421.5 \\
    DDIM \cite{song2020denoising} & 4.67 & 0.53 & 50 & 4.01 \\
    FastDDPM \cite{kong2021fast} & 3.41 & 0.56 & 50 & 4.01 \\
    Recovery EBM \cite{gao2021learning} & 9.58 & - & 180 & - \\
    DDPM Distillation \cite{luhman2021knowledge} & 9.36 & 0.51 & 1 & - \\
    \midrule
    StyleGAN2 w/o ADA \cite{karras2020analyzing} & 8.32 & 0.41 & 1 & 0.04 \\
    StyleGAN2 w/ ADA \cite{karras2020training} & 2.92 & 0.49 & 1 & 0.04 \\
    StyleGAN2 w/ Diffaug \cite{karras2020training} & 5.79 & 0.42 & 1 & 0.04 \\
    \midrule
    Glow \cite{kingma2018glow} & 48.9 & - & 1 & - \\
    PixelCNN \cite{oord2016pixel} & 65.9 & - & 1024 & - \\
    NVAE \cite{vahdat2020nvae} & 23.5 & 0.51 & 1 & 0.36 \\
    VAEBM \cite{xiao2021vaebm} & 12.2 & 0.53 & 16 & 8.79 \\
    \bottomrule
  \end{tabular}
\addtolength{\tabcolsep}{+1pt} 
  }
    \vspace{-2mm}
  \caption{Results on CIFAR-10. Among diffusion models, our method attains a state-of-the-art speed that is close to the speed of StyleGAN2\cite{karras2020training,zhao2020differentiable} counterparts while preserving comparable image fidelity. Since the image resolution is too small (32), we do not use the wavelet-embedded generator in our network on this set. `$^*$' means the speed reproduced on our machine. }
  \label{tab:cifar10}
    \vspace{-2mm}
\end{table}

\minisection{Evaluation metrics} We measure image fidelity by Frechet inception distance (FID)~\cite{heusel2017gans} and measure sample diversity by Recall metric~\cite{kynkaanniemi2019improved}. Following~\cite{jolicoeur2021gotta}, FID and Recall are computed over 50k generated samples. To further demonstrate our faster sampling, we measure the average inference time over 300 trials for a batch size of 100. Besides, the inference time of high-resolution images like CelebA-HQ 512$\times$512 is computed from batches of 25 samples.

\minisection{Implementation details} Our implementation is mainly based on DDGAN~\cite{xiao2021tackling}. We adopt the same training configurations as DDGAN~\cite{xiao2021tackling} for all experiments. Training epochs are set as 500 for CelebA-HQ 256$\times$256, 500 for LSUN 256$\times$256, and 1800 for CIFAR-10 32$\times$32. For CelebA-HQ (512 \& 1024), we train our model and DDGAN for 400 epochs. We train our models on 1 - 8 NVIDIA A100 GPUs for each corresponding dataset. Notably, our models require fewer GPU resources and computations than DDGAN~\cite{xiao2021tackling} thanks to the efficiency of our wavelet diffusion framework as demonstrated in \cref{tab:model_specs}. At the same dataset, our model has a comparable number of parameters but requires less computing FLOPs and memory usage compared with DDGAN~\cite{xiao2021tackling}. Following DDGAN~\cite{xiao2021tackling}, we use 2-sampling steps for CelebA-HQ (256, 512 \& 1024) and 4-sampling steps for CIFAR-10 (32), STL-10 (64), and LSUN-Church (256) in both training and testing.

\begin{table}[t]
  \centering
  \resizebox{\linewidth}{!}{
  \begin{tabular}{l c c c}
    \toprule
    Model & FID$\downarrow$ & Recall$\uparrow$ & Time (s)$\downarrow$ \\
    \midrule
    Ours + W-Generator & \textbf{12.93} & \textbf{0.41} & \textbf{0.38} \\
    DDGAN \cite{xiao2021tackling} & 21.79 & 0.40 & 0.58 \\
    \midrule
    StyleGAN2 w/o \cite{karras2020training} & 11.70 & 0.44 & - \\
    StyleGAN2 w/ ADA \cite{zhao2020differentiable} & 13.72 & 0.36 & - \\
    StyleGAN2 + DiffAug\cite{zhao2020differentiable} & 12.97 & 0.39 & - \\
    \bottomrule
  \end{tabular}
  }
    \vspace{-2mm}
  \caption{Results on STL-10 $64\times64$}
  \label{tab:stl10}
    \vspace{-1mm}
\end{table}

\begin{figure}[t]
  \centering
   \includegraphics[width=0.7\linewidth]{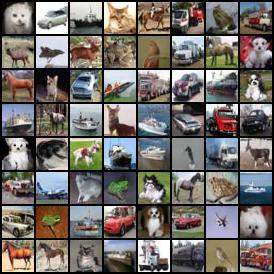}
    \vspace{-2mm}
   \caption{Qualitative results of CIFAR-10 $32\times32$}
   \label{fig:cifar_samples}
    \vspace{-1mm}
\end{figure}

\begin{table}[t]
  \centering
  \begin{tabular}{l c c c}
    \toprule
    Model & FID$\downarrow$ & Recall$\uparrow$ & Time (s)$\downarrow$ \\
    \midrule
    Ours & 6.55 & 0.35 & \textbf{0.60} \\
    Ours + W-Generator & \textbf{5.94} & \textbf{0.37} & 0.79 \\
    DDGAN \cite{xiao2021tackling} & 7.64 & 0.36 & 1.73 \\
    \midrule
    Score SDE \cite{song2020score} & 7.23 & - & - \\
    NVAE \cite{vahdat2020nvae} & 29.7 & - & - \\
    VAEBM \cite{xiao2021vaebm} & 20.4 & - & - \\
    PGGAN \cite{karras2017progressive} & 8.03 & - & - \\
    VQ-GAN \cite{esser2021taming} & 10.2 & - & - \\
    \bottomrule
  \end{tabular}
    \vspace{-2mm}
  \caption{Results on CelebA-HQ $256\times256$. ``Ours'' denotes employing wavelet-based diffusion scheme without using the wavelet-embedded generator.}
  \label{tab:celeba}
\end{table}

\begin{figure}[t]
  \centering
   \includegraphics[width=\linewidth]{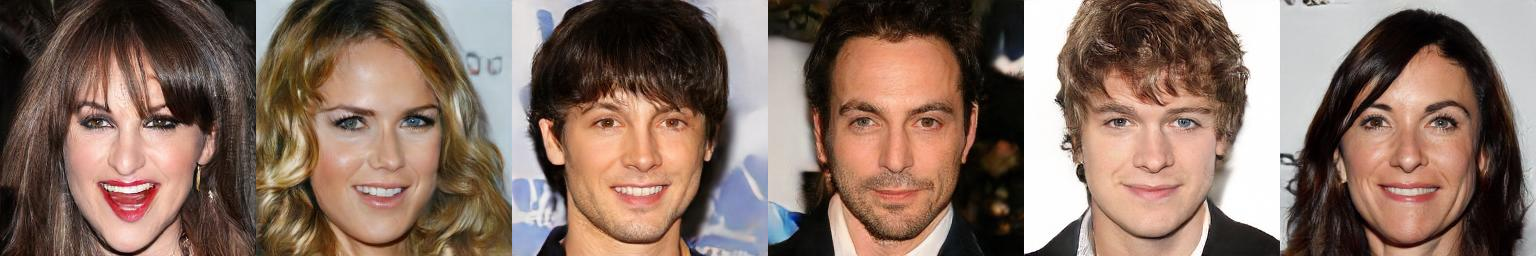}
    \vspace{-6mm}
   \caption{Qualitative results of CelebA-HQ $256\times256$}
   \label{fig:celeb_samples}
    \vspace{-2mm}
\end{figure}

Our speed gain does not only come from the proposed framework but also from corresponding network configurations. As the wavelet input of our model is $4\times$ smaller than the original dimensions, we need to provide suitable network configurations. On CIFAR-10, we use only 3 layers for both the generator and the discriminator instead of 4 as in DDGAN. On CelebA-HQ (256) and LSUN-Church, we use 5 instead of 6 layers. On other datasets, we use the same configurations as in DDGAN, with 6 layers for CelebA-HQ (512 \& 1024) and 4 layers for STL-10.

    

\subsection{Experimental results} 

\minisection{CIFAR-10}
As shown in~\cref{tab:cifar10}, we have greatly improved inference time by requiring only $0.08(s)$ with our Wavelet based diffusion process, which is $2.5\times$ faster than DDGAN. This gives us a real-time performance along with StyleGAN2 methods while exceeding other diffusion models by a wide margin in terms of sampling speed. 

\begin{figure*}[t]
  \centering
   \includegraphics[width=0.75\linewidth]{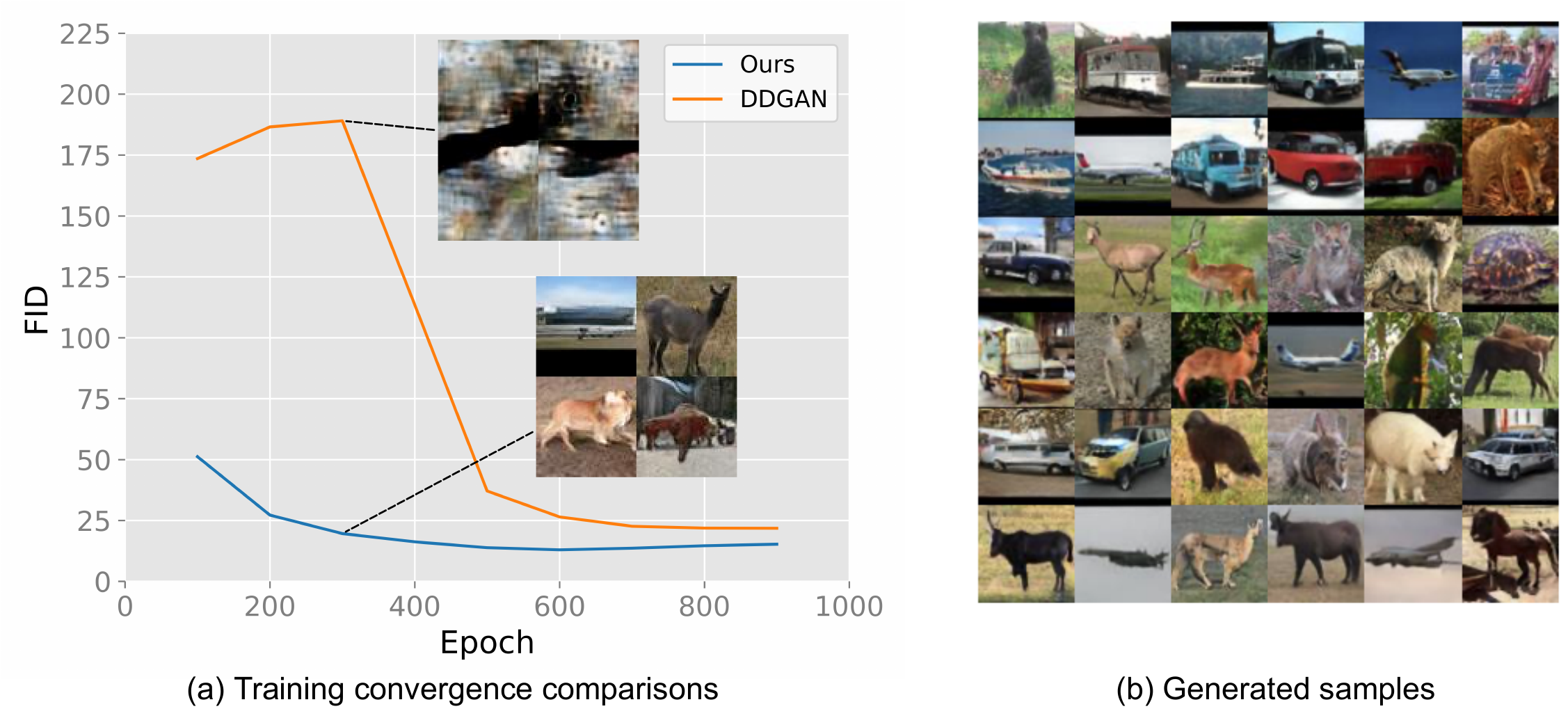}
   \vspace{-3mm}
   \caption{Training convergence comparisons (a) and qualitative results (b) on STL-10 $64\times64$. Our method not only converges faster but also acquires better FID scores than DDGAN~\cite{xiao2021tackling} across different epochs.}
   \label{fig:stl10}
    \vspace{-1mm}
\end{figure*}

\minisection{STL-10}
In~\cref{tab:stl10}, we not only achieve a better FID score at $12.93$ but also gain faster sampling time at $0.38(s)$. We further analyze the convergence speed of our approach and DDGAN in \cref{fig:stl10}a. Our approach offers a faster convergence than the baseline, especially in the early epochs. Generated samples of DDGAN can not recover objects' overall shape and structure during the first 400 epochs. Besides, we provide sample images generated by our model in \cref{fig:stl10}b.

\begin{figure}[t]
  \centering
   \includegraphics[width=0.8\linewidth]{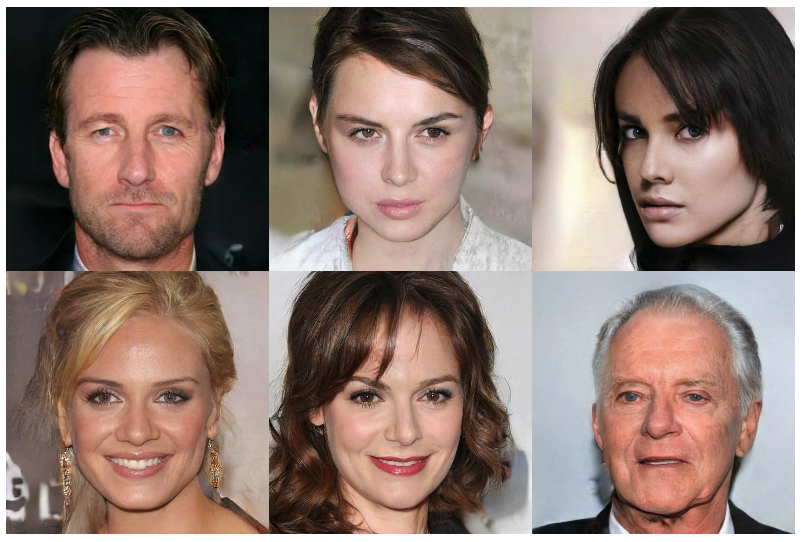}
    \vspace{-3mm}
   \caption{Qualitative results of CelebA-HQ $512\times 512$}
   \label{fig:celeb_samples_512}
    \vspace{-1mm}
\end{figure}

\minisection{CelebA-HQ}
At resolution $256\times256$, we outperform some notable diffusion and GAN baselines with FID at $5.94$ and Recall at $0.37$ while achieving more than $2\times$ faster than DDGAN, as reported in ~\cref{tab:celeba}. On high-resolution CelebA-HQ (512) in~\cref{tab:celeba512}, our model is remarkably better than the DDGAN counterpart for both image quality ($6.40$ vs. $8.43$) and sampling time ($0.59$ vs. $1.49$). We also obtain a higher recall at $0.35$ on high-resolution CelebA-HQ (512). Our further benchmarking on CelebA-HQ 1024 yielded a competitive FID score of $5.98$, comparable to many high-res GAN baselines (StyleGAN - $5.06$, PGGAN - $7.3$). Notably, its inference time (0.59s for 25 samples) remains comparable to CelebA-HQ 512, thanks to our unchanged network configurations with only two modifications: removal of attention layers and use of patch size 2. For qualitative results, we provide the generated samples for the CelebA-HQ (256) and CelebA-HQ (512) datasets in \cref{fig:celeb_samples} and \cref{fig:celeb_samples_512}, respectively. 


\begin{table}[t]
  \centering
    
  \begin{tabular}{l c c c}
    \toprule
    Model & FID$\downarrow$ & Recall$\uparrow$ & Time (s)$\downarrow$ \\
    \midrule
    Ours + W-Generator & \textbf{6.40} & \textbf{0.35} & \textbf{0.59} \\
    DDGAN \cite{xiao2021tackling} & 8.43 & 0.33 & 1.49 \\
    \bottomrule
  \end{tabular}
    \vspace{-2mm}
  \caption{Results on CelebA-HQ $512\times512$}
  \label{tab:celeba512}
    \vspace{-2mm}
\end{table}

\begin{figure}[t]
  \centering
   \includegraphics[width=0.77\linewidth]{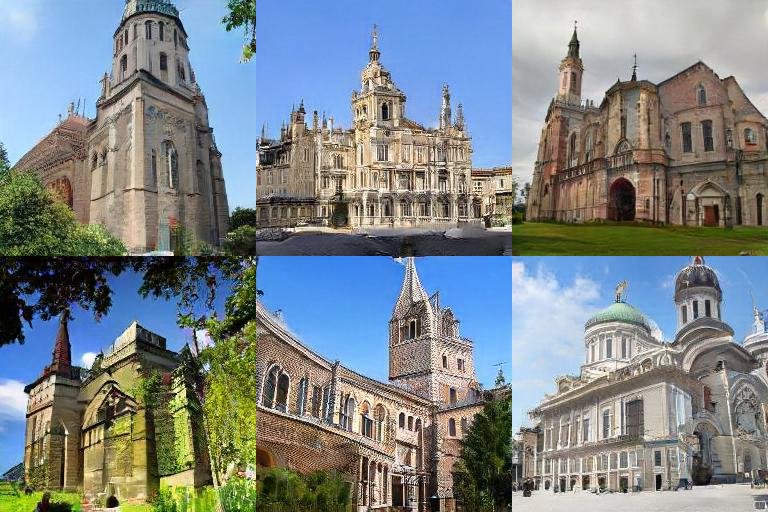}
   \vspace{-2mm}
   \caption{Qualitative results of LSUN-Church $256\times256$.}
   \vspace{-3mm}
   \label{fig:lsun_samples}
\end{figure}

\minisection{LSUN-Church}
In~\cref{tab:lsun}, we obtain a superior image quality at 5.06 in comparison with other diffusion models while achieving comparable results with GAN counterparts. Notably, our model offers $2\times$ faster inference time compared with DDGAN while exceeding StyleGAN2~\cite{karras2020training} in terms of sample diversity at $0.40$. For qualitative results, randomly generated samples are shown at \cref{fig:lsun_samples}.

\subsection{Ablation studies}

\minisection{Reconstruction term} In this section, we verify the contribution of reconstruction loss~\cref{eq:rec_loss} to the model performance on CelebA-HQ $256 \times 256$. The FID score experiences a remarkable reduction of around $0.6$ points to $5.94$ when employing the term, confirming its usefulness in improving model's quality.

\begin{table}[t]
  \centering
    
  \begin{tabular}{l c c c}
    \toprule
    Model & FID$\downarrow$ & Recall$\uparrow$ & Time (s)$\downarrow$ \\
    \midrule
    Ours + W-Generator & \textbf{5.06} & \textbf{0.40} & \textbf{1.54} \\
    DDGAN \cite{xiao2021tackling} & 5.25 & - & 3.42 \\
    \midrule
    DDPM \cite{ho2020denoising} & 7.89 & - & - \\
    ImageBART \cite{esser2021imagebart} & 7.32 & - & - \\
    \midrule
    PGGAN \cite{karras2017progressive} & 6.42 & - & - \\
    StyleGAN \cite{karras2019style} & 4.21 & - & - \\
    StyleGAN2 \cite{karras2020training} & 3.86 &0.36 & - \\
    \bottomrule
  \end{tabular}
    \vspace{-2mm}
  \caption{Results on LSUN-Church $256\times256$.}
  \label{tab:lsun}
  \vspace{-1mm}
\end{table}

\minisection{Wavelet-embedded networks} We validate the contribution of each individual component of our proposed wavelet-based generator on CelebA-HQ $256 \times 256$ in~\cref{tab:ablation_425(500)_wgen}, where the full model includes residual connections, upsampling and downsampling blocks, and bottleneck blocks. As can be seen, each component has a positive effect on the model's performance. By applying all three proposed components, our method achieves the best performance at 5.94, especially the bottleneck block is the least important component in the design of generator. Still, the performance gain comes along with a small cost in running speed. 

\subsection{Running time when generating a single image}

We further demonstrate the superior speed of our models on single images, as expected in real-life applications. In \cref{tab:single_image}, we present their time and key parameters. Our wavelet diffusion models can produce images up to $1024\times1024$ in a mere 0.1s, which is the first time for a diffusion model to achieve such almost real-time performance.

  
    

\begin{table}[t]
  \centering
  \begin{tabular}{l |c c }
    \toprule
    Model & FID$\downarrow$ & Time (s)$\downarrow$ \\
    \midrule
    w/o residual &6.25 & 0.78 \\
    w/o up \& down & 6.23 & \textbf{0.61} \\
    w/o bottleneck & 6.18 & 0.78 \\
    full model &\textbf{5.94} & 0.79 \\
    \bottomrule
  \end{tabular}
   \vspace{-1mm}
  \caption{Ablation study of wavelet generator on CelebA-HQ 256$\times$256. Each setting is trained for 500 epochs.} 
   \vspace{-3mm}
  \label{tab:ablation_425(500)_wgen}
\end{table}

\begin{table}[t]
  \centering
  \resizebox{\linewidth}{!}{
\addtolength{\tabcolsep}{-2.6pt}  
    
  \begin{tabular}{l c c c c c c}
    \toprule
     & CIFAR-10 & STL-10  & CelebA(256) & CelebA(512) & CelebA(1024) & Church \\
    \midrule
    Resolution & 32 & 64 & 256 & 512 & 1024 & 256 \\
    \#time steps & 4 & 4 & 2 & 2 & 2 & 4 \\
    Time (s) & 0.07 & 0.12 & 0.08 & 0.1 & 0.12 & 0.16 \\
    \bottomrule
  \end{tabular}
\addtolength{\tabcolsep}{+2.5pt}  
}
   \vspace{-2mm}
  \caption{Running time when using our full model to generate a single image on each benchmark set.}
  \label{tab:single_image}
\end{table}
    

\subsection{Why ours converges faster and more stably?}
We believe that our model benefits from the frequency decomposition of wavelet transformation. Instead of learning from an entanglement of coarse and detailed information, our method separates them for efficient training and at multi scales in the feature space. First, it studies the low-frequency subbands easier due to lower spatial dimensions. Second, it quickly learns the sparse and repetitive high-frequency components, focusing on distinctive details.

\section{Conclusions} \label{sec:conclusion}
This paper introduced a novel wavelet-based diffusion scheme that demonstrates superior performance on both image fidelity and sampling speed. By incorporating wavelet transformations to both image and feature space, our method can achieve the state-of-the-art running speed for a diffusion model, closing the gap with StyleGAN models\cite{karras2019style,karras2020training,zhao2020differentiable} while obtaining a comparable image generation quality to StyleGAN2 and other diffusion models. Besides, our method offers a faster convergence than the baseline DDGAN~\cite{xiao2021tackling}, confirming the efficiency of our proposed framework. With these initial results, we hope our approach can facilitate future studies on real-time and high-fidelity diffusion models.

{\small
\bibliographystyle{ieee_fullname}
\bibliography{egbib}
}

\newpage
\appendix
\section{Sensitivity analysis of training batch size}\label{apsec:sensitivity}

\begin{figure}[t]
  \centering
   \begin{subfigure}[b]{0.85\linewidth}
         \centering
         \includegraphics[width=\linewidth]{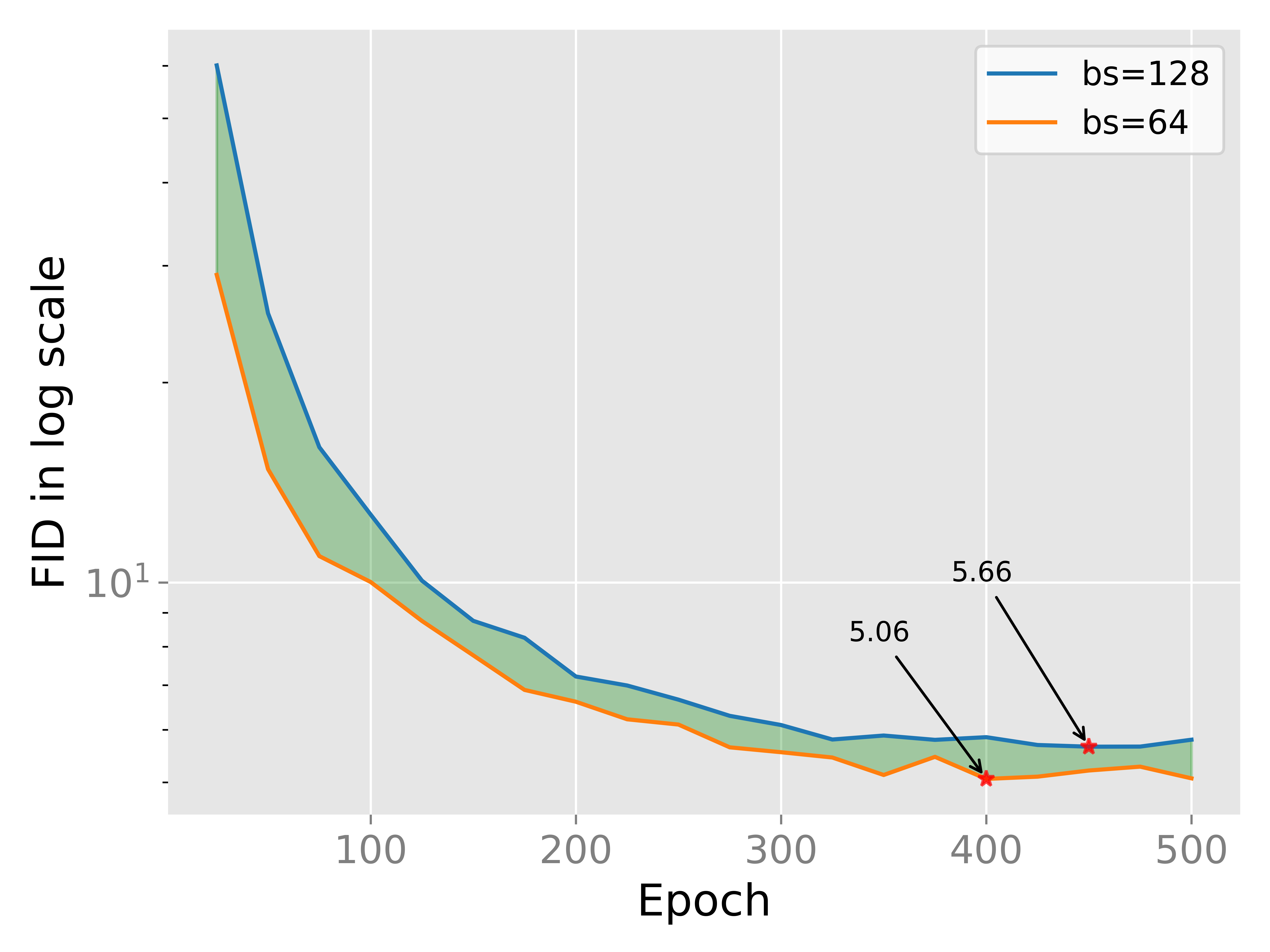}
         \caption{LSUN-Church}
         \label{fig:lsun_curves}
   \end{subfigure}
   \hfill
   \begin{subfigure}[b]{0.85\linewidth}
         \centering
         \includegraphics[width=\linewidth]{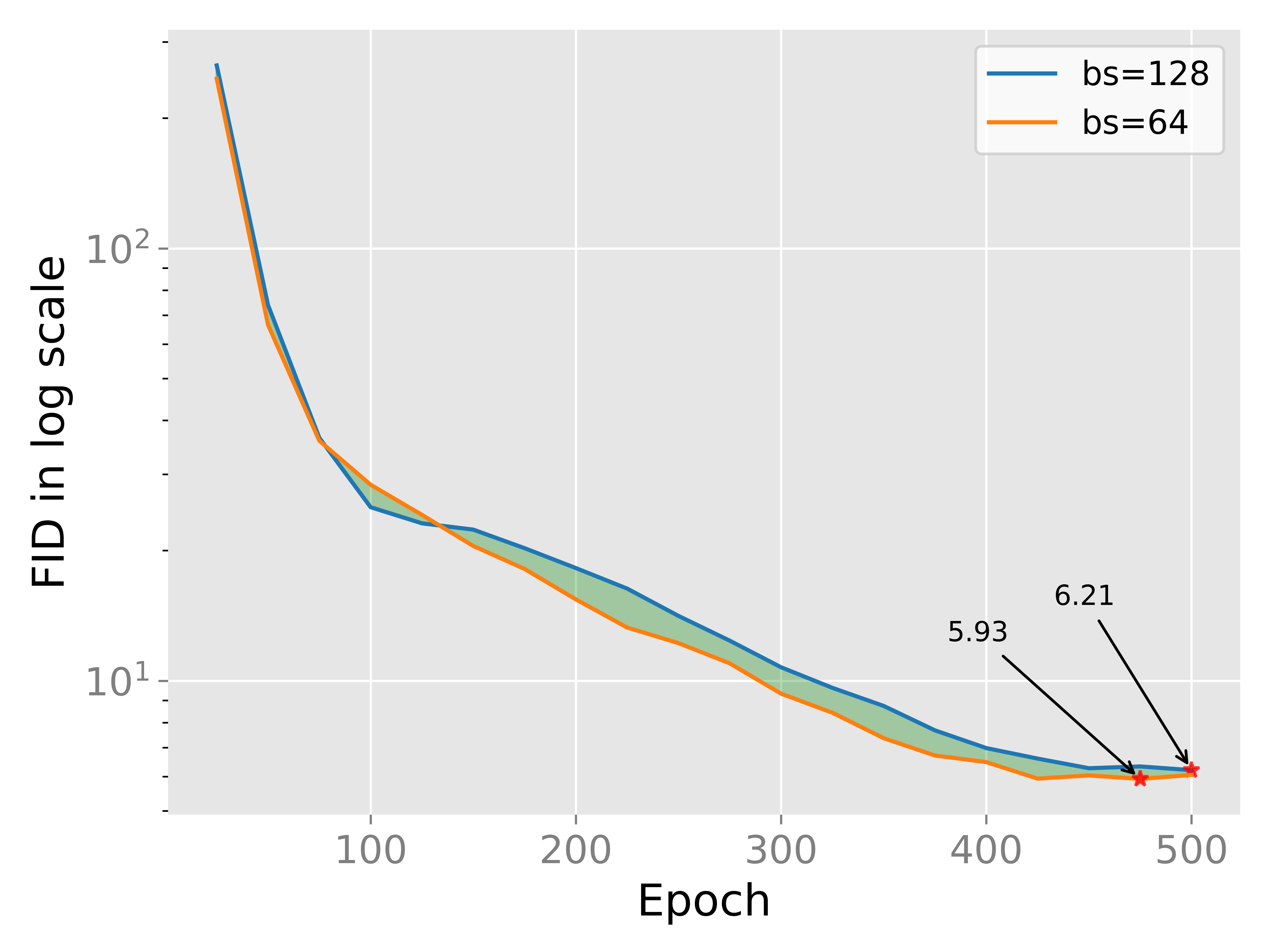}
         \caption{CelebA-HQ}
         \label{fig:cel_curves}
   \end{subfigure}

   \caption{Training curves on LSUN-Church and CelebA-HQ ($256\times256$) with two different batch sizes, namely 64 and 128. The area under curves (\text{\color{green}green color}) presents the difference in FID scores between the two trained models. Unlike CelebA-HQ (256), the gap between the two models on LSUN-Church is notably large across different epochs. By using an appropriate batch size of 64, the minimum FID of the trained model can reduce to 5.06. On CelebA-HQ (256), it again confirms that using sufficient batch size is highly necessary for the model performance as the model is trained with a batch size of 64 attaining a better FID score of 5.93.}
   \label{fig:training_curves}
\end{figure}

We recognize that training batch size is a critical aspect affecting the final performance. Large batch size often results in worse performance, with the compensation being training time. Here, we carefully analyze the effect of training batch size on the model performance measured by Frechet inception distance (FID)~\cite{heusel2017gans} (depicted in \cref{fig:training_curves}).

As expected, the model trained with batch size 64 consistently performs better than the model trained with batch size 128, as illustrated via the training curves on LSUN-Church in \cref{fig:lsun_curves}. The gap is initially large during the first 100 epochs and then shrunk in the following epochs. However, the performance gap is still significant. The best FID of the model with batch size 64 is 5.06, which is 0.6 points lower than the best FID of the model trained on batch size 128. More importantly, our model outperforms DDGAN (5.06 vs. 5.25) when using the same batch size of 64.

\begin{table*}[t]
\centering

\begin{tabular}{l|c|c|c|c|c}
\toprule
                                  & CIFAR-10 & STL-10    & CelebA-HQ (256) & CelebA-HQ (512) & LSUN-Church \\
\midrule
\# of ResNet blocks per scale     & 2        & 2         & 2               & 2               & 2           \\
Base channels                    & 128      & 128       & 64              & 64              & 64          \\
Channel multiplier per scale & (1,2,2)  & (1,2,2,2) & (1,2,2,2,4)     & (1,1,2,2,4,4)   & (1,2,2,2,4) \\
Attention resolutions              & None        & 16        & 16              & 16              & 16          \\
Latent Dimension                  & 100      & 100       & 100             & 100             & 100         \\
\# of latent mapping layers       & 4        & 4         & 4               & 4               & 4           \\
Latent embedding dimension        & 256      & 256       & 256             & 256             & 256    
     \\

\bottomrule
\end{tabular}

\caption{Network configurations. }
\label{tab:net_config}
\end{table*}

We further verify the effect of batch size to our trained models on CelebA-HQ (256) with batch sizes 128 and 64. As shown in \cref{fig:cel_curves}, the model trained with batch size 64 achieves a minimum FID of 5.93, which is $0.28$ points lower than the FID 6.21 of the model trained with batch size 128. It again confirms that batch size is an important factor to be considered when evaluating and comparing model performance.

Note that we used a larger batch size than DDGAN (32 vs. 16) on CelebA-HQ $512\times512$ with 8 GPUs\footnote{NVIDIA A100-40GB GPUs are used. Our model can fit a training batch size of 4 instead of 2 as DDGAN per GPU.}. Due to time and resource limits, we cannot retrain our model with batch size 16, but we expect our result will be further improved in this fair experiment configuration, further increasing the gap in performance between our algorithm and DDGAN.

\section{Experimental details} \label{apsec:exp_details}

\subsection{Wavelet transformations} We utilize the implementation of wavelet transformations, including Discrete wavelet transform (DWT) and Discrete inverse wavelet transform (IWT), from \cite{li2020wavelet}. We perform these transformations on both input images and feature maps for further processing in the proposed Wavelet-based Diffusion framework.



\subsection{Network configurations}
\minisection{Generator.} Our generator has a UNet alike architecture\cite{ronneberger2015u} which is mainly based on NCSN++ \cite{song2020score,xiao2021tackling}. As can be seen in \cref{tab:net_config}, we show the detailed configurations of the generator for each corresponding dataset. We adjust the number of layers in the generator according to the input resolution of wavelet coefficients. The number of channels of time embedding is $4\times$ larger than the base channels.

\minisection{Discriminator.} The number of layers in the discriminator is the same as the one of the generator. For more details of the discriminator structure, please refer to \cite{xiao2021tackling}.


\subsection{Training hyper-parameters}
For reproductivity, we further provide a full table of tuned hyper-parameters in \cref{tab:hyperparams}. Basically, our hyper-parameters are the same as the baseline \cite{xiao2021tackling} except for the number of epochs and the allocated GPUs on specific datasets. Meanwhile, there are two new datasets, including STL-10 $64\times64$ and CelebA-HQ $512 \times 512$, that share similar configurations of CIFAR-10 and CelebA-HQ $256 \times 256$, respectively. Besides, the setting of CelebA-HQ $1024 \times 1024$ is almost similar to CelebA-HQ $512 \times 512$. 

For training time, CIFAR10 and STL10 models require 1.6 and 3.6 days on a single GPU, respectively. On CelebA-HQ 256 and LSUN-Church, they take 1.1 and 6.8 days on 2 and 4 GPUs, respectively. On high-resolution CelebA-HQ 512, it takes 4.3 days on 8 GPUs. Besides, the training time is mainly influenced by the number of denoising steps, the size of network architectures, and image resolutions (presented in \cref{tab:net_config}) apart from the number of training epochs. This is also the same for the inference time.

\begin{table*}[t]
    \centering
    \begin{tabular}{l|c|c|c|c|c}
        \toprule
                                            & CIFAR-10  & STL-10    & CelebA-HQ (256 \& 512) & LSUN-Church & CelebA-HQ (1024) \\
        \midrule
        $\text{lr}_{G}$                            & $1.6\text{e-4}$  & $1.6\text{e-}4$  & $2.\text{e-}4$      & $1.6\text{e-}4$ & $2.\text{e-}4$   \\
        $\text{lr}_{D}$                            & $1.25\text{e-}4$ & $1.25\text{e-}4$ & $1.\text{e-}4$      & $1.\text{e-}4$ & $1.\text{e-}4$    \\
        Adam optimizer ($\beta_1$ \& $\beta_2$) & 0.5, 0.9  & 0.5, 0.9  & 0.5, 0.9 & 0.5, 0.9 & 0.5, 0.9  \\
        EMA                                 & 0.9999    & 0.9999    & 0.999       & 0.999 & 0.999     \\
        Batch size                          & 256       & 256       & 64 \& 32       & 128 & 24     \\
        Lazy regularization                 & 15        & 15        & 10          & 10 & 10        \\
        \# of epochs                        & 1800      & 900       & 500 \& 400     & 500 & 400     \\
        \# of timesteps                     & 4         & 4         & 2           & 4 & 2          \\
        \# of GPUs                          & 1         & 1         & 2 \& 8           & 4 & 8    \\
        \bottomrule
    \end{tabular}
    \caption{Choices of hyper-parameters}
    \label{tab:hyperparams}
\end{table*}

\section{More qualitative results} \label{apsec:qualitative}
\begin{figure*}[t]
  \centering
   \includegraphics[width=0.8\linewidth]{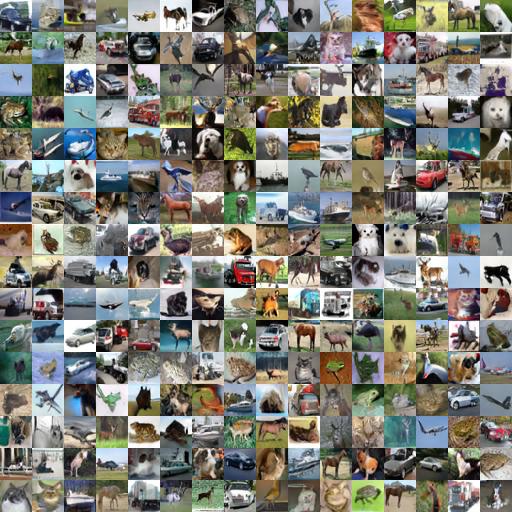}

   \caption{Non-curated generated samples on CIFAR-10}
   \label{fig:add_cifar10_samples}
\end{figure*}

\begin{figure*}[t]
  \centering

   \begin{subfigure}[b]{0.7\linewidth}
         \centering
         \includegraphics[width=\linewidth]{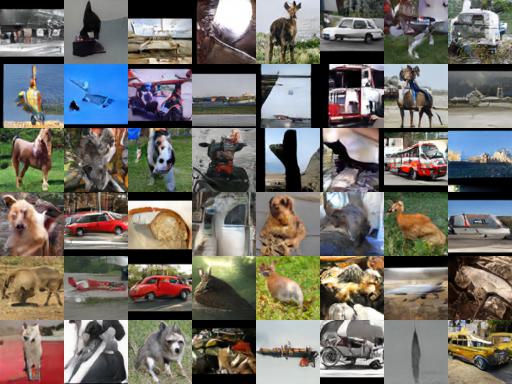}
         \caption{Ours}
         \label{fig:add_stl10_samples}
   \end{subfigure}
   \hfill
   \begin{subfigure}[b]{0.7\linewidth}
         \centering
         \includegraphics[width=\linewidth]{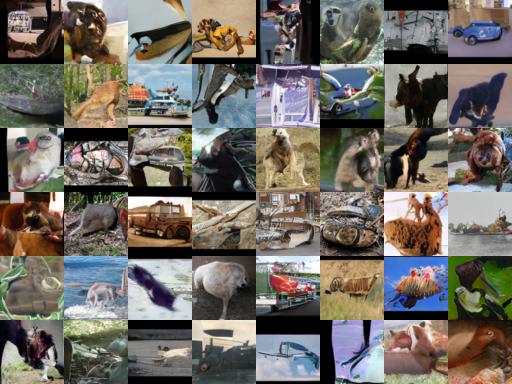}
         \caption{DDGAN}
         \label{fig:add_ddgan_stl10_samples}
   \end{subfigure}

   \caption{Non-curated generated samples between ours and DDGAN\cite{xiao2021tackling} on STL-10}
   \label{fig:comparisons_stl10_samples}
\end{figure*}

\begin{figure*}[t]
  \centering
   \includegraphics[width=0.8\linewidth]{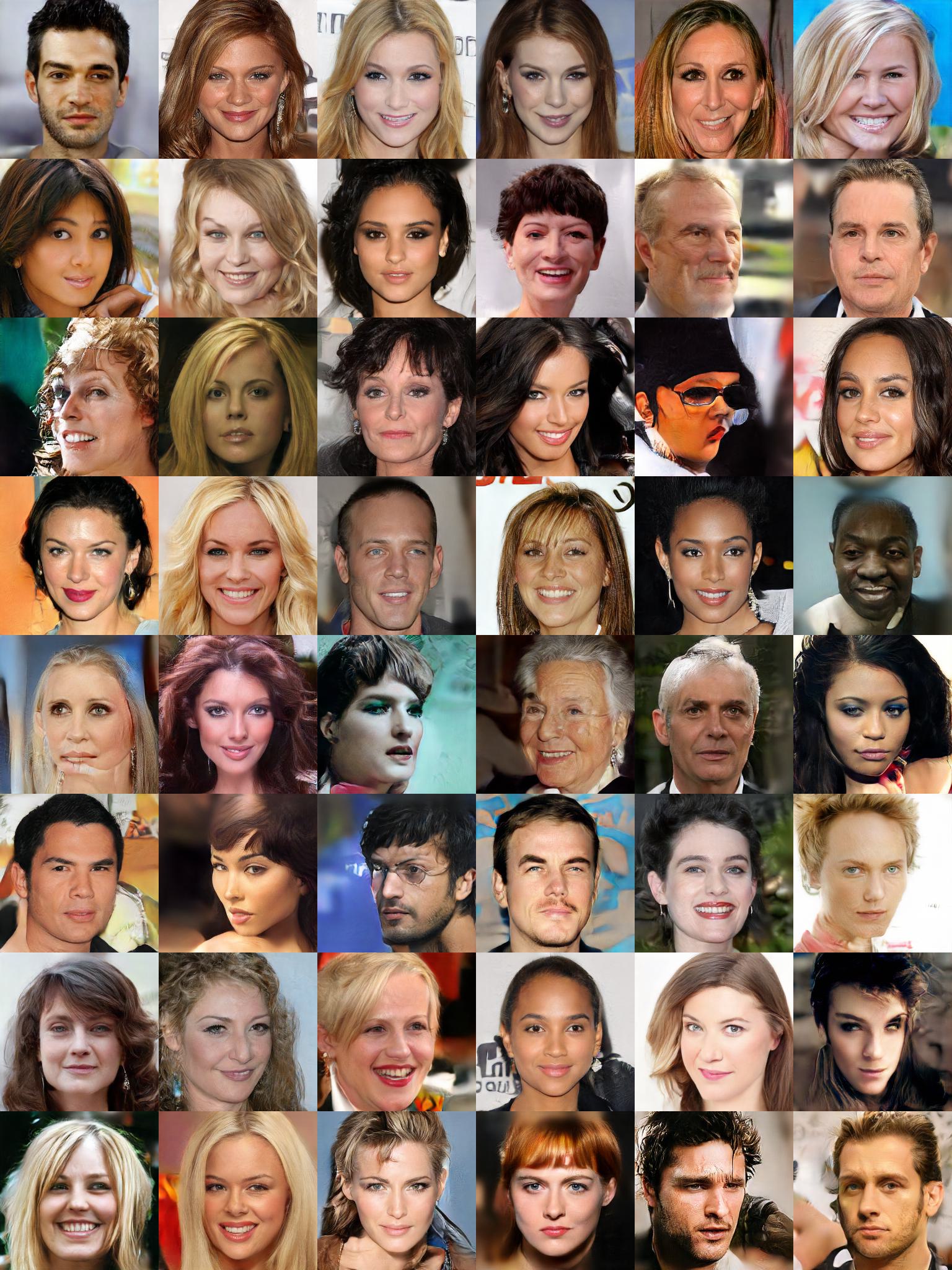}

   \caption{Non-curated generated samples on CelebA-HQ $256 \times 256$}
   \label{fig:add_cel256_samples}
\end{figure*}

\begin{figure*}[t]
  \centering
   \includegraphics[width=0.6\linewidth]{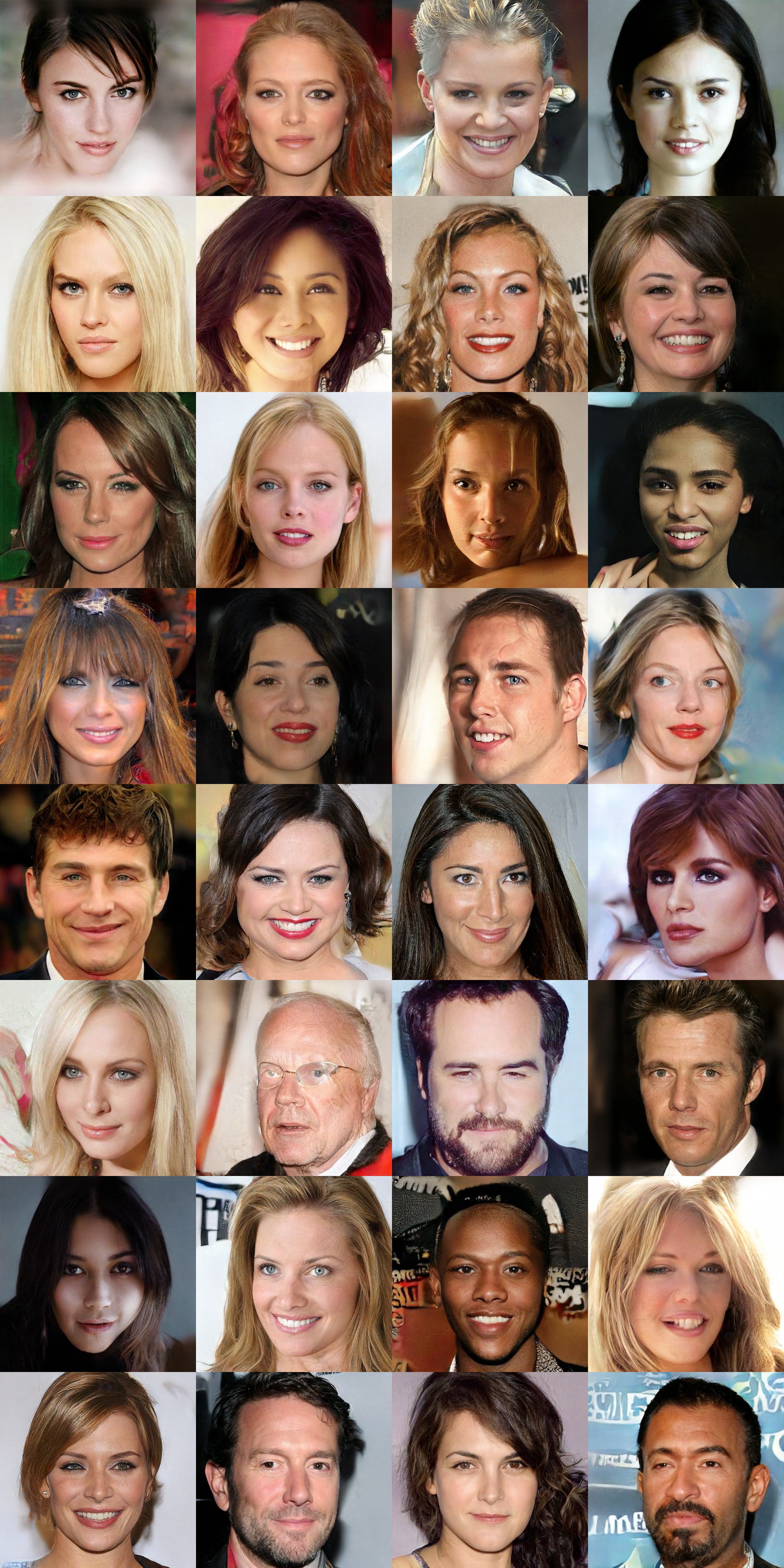}

   \caption{Non-curated generated samples on CelebA-HQ $512 \times 512$}
   \label{fig:add_cel512_samples}
\end{figure*}

\begin{figure*}[t]
  \centering
   \includegraphics[width=0.8\linewidth]{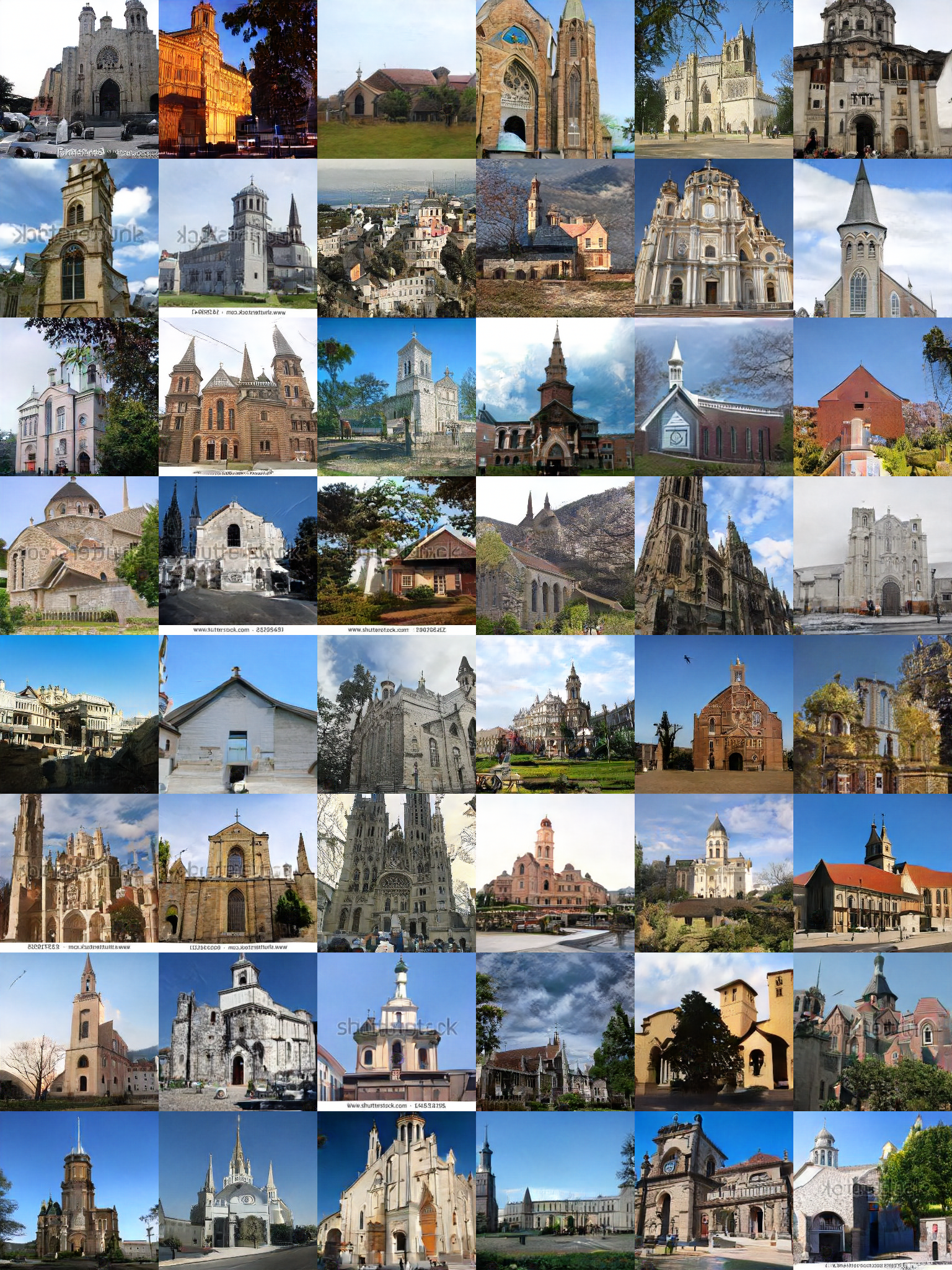}

   \caption{Non-curated generated samples on LSUN-Church}
   \label{fig:add_lsun_samples}
\end{figure*}

\begin{figure*}[t]
  \centering
   \includegraphics[width=0.6\linewidth]{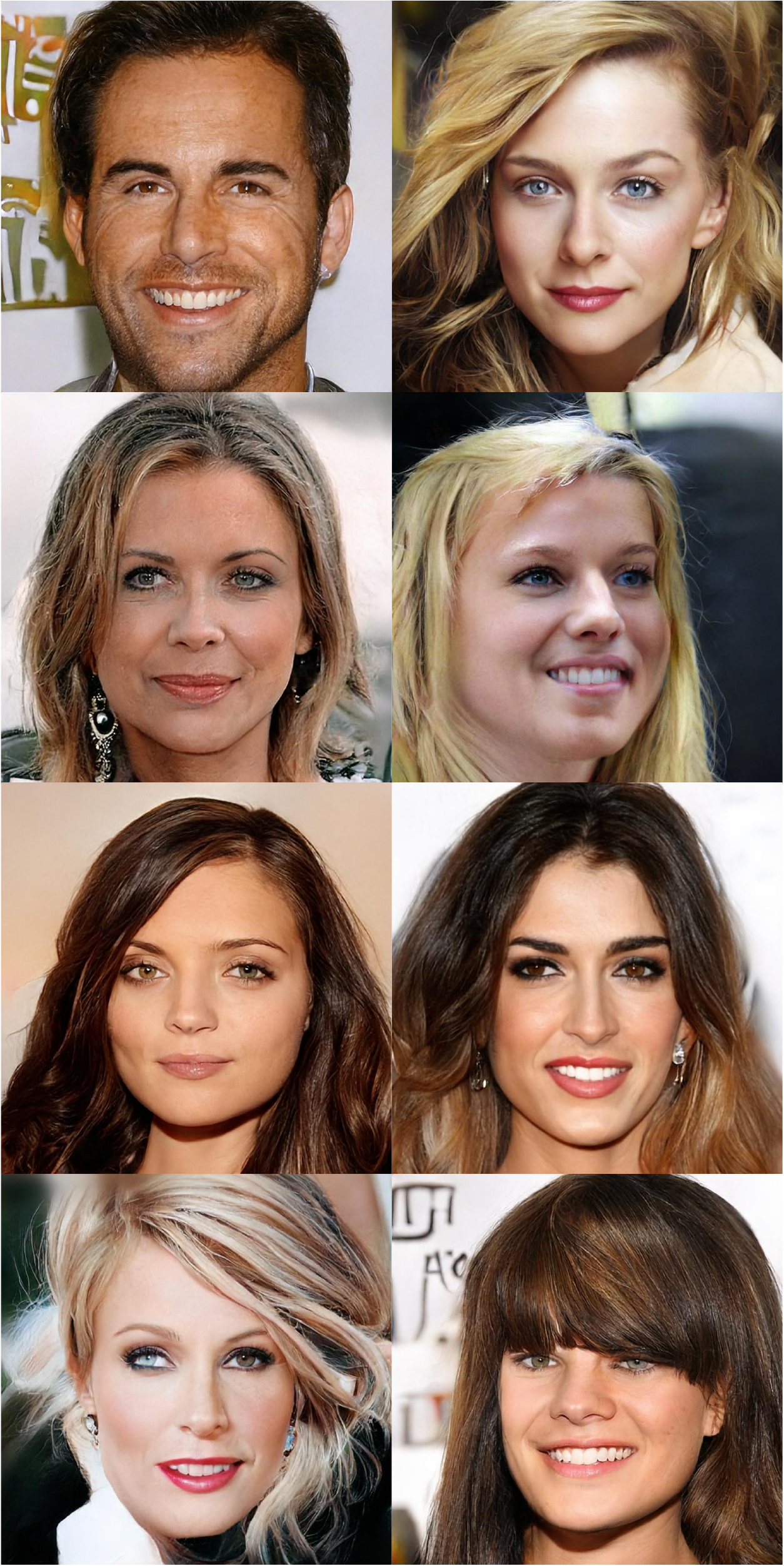}

   \caption{Non-curated generated samples on CelebA-HQ $1024 \times 1024$}
   \label{fig:add_cel1024_samples}
\end{figure*}

We further provide additional qualitative results on CIFAR-10 in \cref{fig:add_cifar10_samples}, STL-10 in \cref{fig:add_stl10_samples}, CelebA-HQ 256 in \cref{fig:add_cel256_samples}, CelebA-HQ 512 in \cref{fig:add_cel512_samples}, LSUN-Church in \cref{fig:add_lsun_samples}, and CelebA-HQ 1024 in \cref{fig:add_cel1024_samples}.

\begin{figure*}[t]
    \centering
    \includegraphics[width=.8\linewidth]{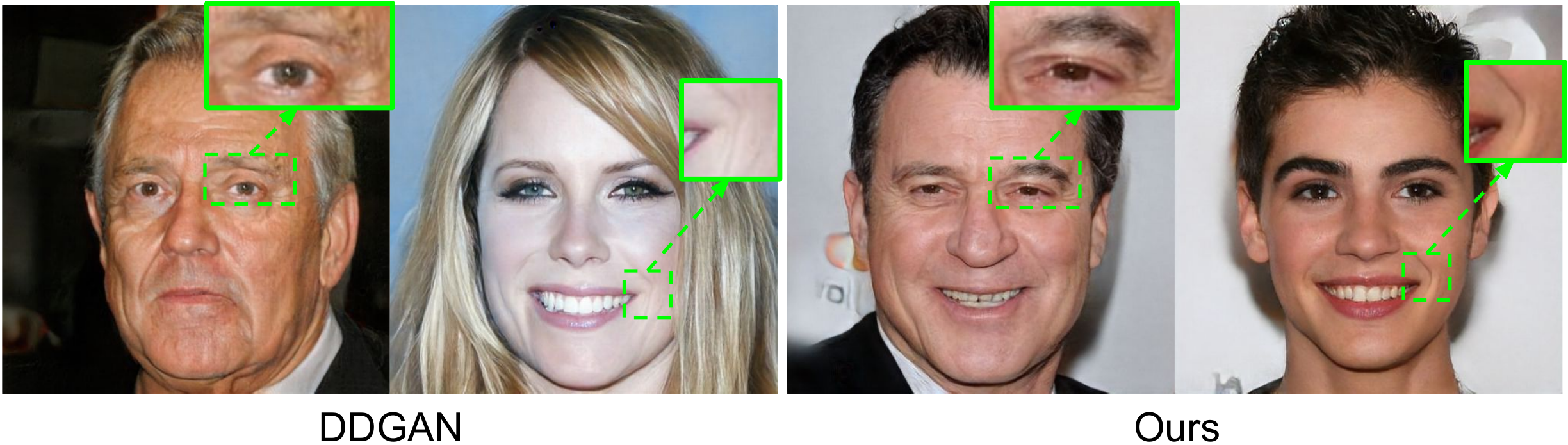}

    \caption{Qualitative comparision on CelebA-HQ $512 \times 512$}
    \label{fig:comparison_cel512_samples}
    \vspace{-5mm}
\end{figure*}

A comparison of qualitative samples between ours and DDGAN\cite{xiao2021tackling} on STL-10 is also presented in \cref{fig:comparisons_stl10_samples}. Our model clearly achieves better sample quality with a plausible appearance of generated objects, while the counterpart fails to represent object-specific shapes in output samples. We also add a qualitative comparison on the CelebA-HQ 512 dataset (\cref{fig:comparison_cel512_samples}), which further illustrates the advantages of our proposal in producing clearer details, such as eyebrows and wrinkles.

\section{More discussion}
\minisection{Connection between wavelet transformation and speed.}
Let $X \in \mathbb{R}^{C\times H\times W}$ denote an input image. Wavelet transformation reduces its spatial dimension while increasing its channel dimension by four-folds: $Y \in \mathbb{R}^{4C\times H/2\times W/2}$. This input is then \textbf{\textit{projected to the base channel D}} via the first linear layer, keeping the network width unchanged compared with DDGAN. Hence, most of the network benefits from $4\times$ reduction in spatial dimensions, significantly reducing its computation (measured in FLOPs). For further acceleration, we decrease the network depth since a smaller spatial dimension requires fewer downsampling steps.

\minisection{Increment novelty.}
While wavelet transformation has been used in many previous works on different tasks, ours is the first work employing it in diffusion models and in a comprehensive manner. Wavelet transformation is carefully incorporated on both the pixel and feature levels. Particularly, for the feature level, we proposed three wavelet-based network components, and each component is designed to utilize low and high-frequency subbands for improved output quality. Thanks to these proposals, we achieve state-of-the-art running speed with a near real-time performance for a diffusion model, allowing this advanced technique to be applicable to real-time applications. Our method also provides a faster and more stable model training, as discussed in Sec 5.5. of the main paper. Hence, we believe our paper is essential and not just an incremental work.

\minisection{Reasons why the high-frequency subbands are unprocessed and directly transmitted to the decoder in the frequency bottleneck block.} 
When designing this block, we aimed to strengthen our model's focus on learning low-level features while preserving the details; hence we directly transmitted the high-frequency components to the IWT module. We have tested processing both low and high subbands on STL-10 and got almost the same FID ($12.96$ vs. $12.93$), suggesting that this design is not critical.

\minisection{Progressive upsampling.}
We actually tried this direction first, but it produced quite poor results. On CelebA-HQ 256, the FID score of the 2-level upsampling network is $13.11$, much higher than ours ($5.94$). We suspect a discrepancy in conditional distribution between low and high-frequency $p(x_{hi}|x_{lo})$, causing mismatching between generated subbands, and deteriorating the output quality.

\end{document}